\documentclass[10pt,twocolumn,letterpaper]{article}

\usepackage{wacv}              

\usepackage{times}
\usepackage{epsfig}
\usepackage{graphicx}
\usepackage{amsmath}
\usepackage{amssymb}
\usepackage{booktabs}
\usepackage{multirow}
\usepackage{subcaption}
\usepackage{tablefootnote}
\usepackage{balance}
\usepackage[accsupp]{axessibility}  

\usepackage[pagebackref,breaklinks,colorlinks]{hyperref}

\usepackage[capitalize]{cleveref}
\crefname{section}{Sec.}{Secs.}
\Crefname{section}{Section}{Sections}
\Crefname{table}{Table}{Tables}
\crefname{table}{Tab.}{Tabs.}

\def\wacvPaperID{11} 
\def\confName{WACV}
\def\confYear{2024}

\begin{document}

\title{Investigating Weight-Perturbed Deep Neural Networks With Application in Iris Presentation Attack Detection}

\author{Renu Sharma, Redwan Sony, Arun Ross\\
Michigan State University\\
{\tt\small \{sharma90, sonymd, rossarun\}@msu.edu}}

\maketitle
\thispagestyle{empty}

\begin{abstract}
Deep neural networks (DNNs) exhibit superior performance in various machine learning tasks, e.g., image classification, speech recognition, biometric recognition, object detection, etc. However, it is essential to analyze their sensitivity to parameter perturbations before deploying them in real-world applications. In this work, we assess the sensitivity of DNNs against perturbations to their weight and bias parameters. The sensitivity analysis involves three DNN architectures (VGG, ResNet, and DenseNet), three types of parameter perturbations (Gaussian noise, weight zeroing, and weight scaling), and two settings (entire network and layer-wise). We perform experiments in the context of iris presentation attack detection and evaluate on two publicly available datasets: LivDet-Iris-2017 and LivDet-Iris-2020. Based on the sensitivity analysis, we propose improved models simply by perturbing parameters of the network without undergoing training. We further combine these perturbed models at the score-level and at the parameter-level to improve the performance over the original model. The ensemble at the parameter-level shows an average improvement of 43.58\% on the LivDet-Iris-2017 dataset and 9.25\% on the LivDet-Iris-2020 dataset. The source code is available at \href{https://github.com/redwankarimsony/WeightPerturbation-MSU}{https://github.com/redwankarimsony/WeightPerturbation-MSU}.
\end{abstract}

\section{Introduction}
\label{sec:introduction}

Deep Neural Networks (DNNs) have revolutionized the machine learning field through their superior performance in various tasks especially in the field of computer vision \cite{ Simonyan2015, He2016, Huang2017}, natural language processing \cite{Deng2018}, and speech technology \cite{Deng2013}. In essence, a DNN comprises a sequence of layers containing trainable parameters (weights and bias) to learn a complex mapping between input signals and output labels. For deploying DNNs in real-world applications, it is crucial to analyze their robustness or sensitivity to hardware/sensor noise introduction \cite{Cheney2017}, environment changes \cite{Tsai2021} and adversarial attacks \cite{Garg2020}. Sensitivity analysis also helps in building a quantized-weights model with commensurate performance \cite{Han2015, Weng2020}.

In the literature, sensitivity analysis of DNNs has been performed by perturbing either the input signal or the architectural parameters. The work in \cite{Szegedy2014, Goodfellow2015, Fawzi2016, Karianakis2016, Moosavi-Dezfooli2017, Novak2018} analyze DNN robustness by manipulating the input signals, whereas the work in \cite{Han2015, Xiang2019, Shu2019, Tsai2021, Weng2020} perturb architectural parameters to analyze robustness. Yeung \textit{et al.} \cite{Yeung2009} provide a detailed sensitivity analysis of neural networks over input and parameter perturbations. In this work, we focus on the sensitivity analysis of DNNs when architectural parameters (learned weights) are perturbed.

The authors in \cite{Shu2019, Xiang2019, Weng2020, Tsai2021} provide a theoretical sensitivity analysis based on parameter perturbations. Shu and Zhu \cite{Shu2019} propose an influence measure motivated by information geometry to quantify the effects of various perturbations to input signals and network parameters on DNN classifiers. Xiang \textit{et al.} \cite{Xiang2019} design an iterative algorithm to compute the sensitivity of a DNN layer by layer, where sensitivity is defined as ``the mathematical expectation of absolute output variation due to weight perturbation with respect to all possible inputs" \cite{Xiang2019}. Tsai \textit{et al.} \cite{Tsai2021} study the robustness of the pairwise class margin function against weight perturbations. Weng \textit{et al.} \cite{Weng2020} compute a certified robustness bound for weight perturbations, within which a neural network will not make erroneous outputs. In addition, they also identify a useful connection between the developed certification and the challenge of weight quantization.

In this work, we empirically analyze the sensitivity of DNNs by manipulating their architectural parameters. We examine sensitivity of three widely used architectures (VGG \cite{Simonyan2015}, ResNet \cite{He2016}, and DenseNet \cite{Huang2017}) under three types of parameter perturbations (Gaussian noise, weight zeroing and weight scaling). We apply the perturbations in two settings: over all the layers of a network simultaneously and over each layer at a time. Our work is motivated from \cite{Cheney2017}, where they also empirically analyze the sensitivity of the pre-trained AlexNet and VGG16 networks to internal architecture and weight perturbations. However, our work is vastly different. Being motivated by their analysis, not only do we analyze the robustness or sensitivity of the newer networks, we also improve those models with different perturbation methods without any training. First, we extend the work by evaluating the sensitivity of heavily used CNN architectures in biometric tasks: VGG, ResNet, and DenseNet. Second, we perform additional weight manipulations (weight scaling, variants of weight zeroing, and additional setting of applying perturbations over the entire network parameters) in the sensitivity analysis. Third, we leverage the findings from the sensitivity analysis and propose an ensemble of perturbed models to improve the performance without any further training. Our main contributions are as follows:

\noindent 1. We perform sensitivity analysis of three DNN architectures (VGG \cite{Simonyan2015}, ResNet \cite{He2016} and DenseNet \cite{Huang2017}) against parameter perturbations.

\noindent 2. We apply a number of parameter perturbations (three types of perturbations and its variant in two settings) to analyze the sensitivity of deep neural networks in the context of iris presentation attack detection.

\noindent 3. We leverage the sensitivity analysis to propose a better performing model by ensembling the perturbed models at two different levels: score-level and parameter-level.

\noindent 4. We perform experiments using five datasets. Three of the datasets (IARPA, NDCLD-2015, Warsaw Postmortem v3) are used for training, whereas the others (LivDet-Iris-2017 and LivDet-Iris-2020) are used for testing. This represents a cross-dataset scenario, where training and testing are performed on different datasets.

The rest of the paper is organized as follows: Section \ref{sec:ParameterPerturbations} provides the details of various parameter perturbations used for the sensitivity analysis of DNNs; Section \ref{sec:ApplicationScenario} describes the application scenario considered in this work; Section \ref{sec:DatasetSetup} explains the dataset and experimental setup; Section \ref{sec:SensitivityAnalysis} provides the sensitivity analysis of the three architectures against the considered parameter perturbations; and Section \ref{sec:ProposedImprovement} describes how we leverage the sensitivity analysis to generate an ensemble of perturbed models for improving performance. Finally, Section \ref{sec:Summary} summarizes the paper and provides future directions.

\section{Parameter Perturbations}
\label{sec:ParameterPerturbations}
We explore the sensitivity of neural networks by perturbing their architectural parameters (weights and bias). From here on, we use the terms `architectural parameters', `parameters', and `weights' interchangeably. 
To measure the sensitivity, we consider the change in the performance of the DNN when weights are perturbed. Let $n$ input samples be $\{x_1, x_2, ..., x_n\}$ and their output be $\{y_1, y_2, ..., y_n\}$. Here, we labeled the positive class as `1' and the negative class as `0'. The predicted output values from a DNN approximator are $\{f(x_1,W_{org}), f(x_2,W_{org}), ..., f(x_n,W_{org})\}$, where $W_{org}$ are the learned parameters. We measure the performance of the DNN in terms of True Detection Rate (TDR). TDR is a percentage of positive samples correctly classified:
\begin{equation}
    TDR_{org} = \frac{\sum_i^n (f(x_i,W_{org}) > T)}{ \sum_i^n y_i} *100 
\end{equation}
where, $T$ is the threshold. The input sample with a predicted value above the threshold is considered a positive class. After weight perturbation, we estimate the output as $\{ f(x_1,W_{mod}), f(x_2,W_{mod}), ..., f(x_n,W_{mod})\}$, where $W_{mod}$ are the perturbed parameters. We then use these predicted values to measure the performance of DNN ($TDR_{mod}$). The higher the change in the performance ($|TDR_{org}- TDR_{mod}|$), the higher the sensitivity of the neural network to the particular perturbation. 

We perturb the parameters in two settings: manipulating parameters of all layers simultaneously and manipulating parameters one layer at a time. The first setting aims to understand the overall sensitivity of DNNs, whereas the second setting examines which layer has more impact on the model. The higher the sensitivity, the lower the generalization of the DNN \cite{Novak2018,Tsai2021}. The three perturbations we consider are Gaussian noise manipulation, weight zeroing, and weight scaling. These perturbations resemble (a) noise introduction due to defects in hardware implementations of neural networks \cite{Mead1989}, and (b) adversarial weight perturbations \cite{Garg2020, Rakin2021} on open-sourced models. Eventually, the choice of perturbations is based on their simplicity. This work has also the potential of obtaining quantized or compressed DNN models, which consume less memory with equivalent performance. Details of these perturbations are as follows:

\noindent \textbf{1. Gaussian Noise Manipulation:}
Here, we manipulate the original parameters of the layers by adding Gaussian noise sampled from a normal distribution of zero mean and scaled standard deviation. We control the scaling of the standard deviation by the scalar factor $\boldsymbol{\alpha}$. The modified parameters are defined as 
\begin{equation}
    W_{mod} = W_{org}+N(0, \boldsymbol{\alpha}*\sigma(W_{org})).
\end{equation}

Here, $W_{org}$ are the original parameters, $W_{mod}$ are the modified parameters, and $N(\mu, \sigma)$ is the normal distribution. We calculate $\sigma(W_{org})$ for a particular layer by first flattening the parameter tensor to a 1-D array and then computing the standard deviation. So, the standard deviation and the Gaussian noise distribution will differ for each layer since $\sigma(W_{org})$ varies from layer to layer. Consequently, the \textit{absolute} perturbations differ for each layer. However, \textit{relative} perturbations are the same across layers.


\noindent \textbf{2. Weight Zeroing:}
In the second manipulation, we randomly select a certain proportion of parameters and set them to zero. The portion of parameters is determined by a scalar factor $\boldsymbol{\beta}$. The modified parameters are represented as 
\begin{equation}
\begin{array}{l}
    W_{mod}[random(\boldsymbol{\beta}, W_{org})] = 0.
\end{array}
\end{equation}

Here, $random(.,.)$ is the function that returns the index of $\boldsymbol{\beta}$ proportion of randomly selected parameters from the original set of parameters. We also define another version of weight zeroing, where weights are first sorted, and then $\boldsymbol{\beta}$ proportion of low-magnitude weights is set to zero. 

\noindent \textbf{3. Weight Scaling:}
The third perturbation scales the original parameters by a scalar factor $\boldsymbol{\gamma}$ as   
\begin{equation}
    W_{mod} = \boldsymbol{\gamma} * W_{org}.
\end{equation}


\section{Application Scenario}
\label{sec:ApplicationScenario}
We perform sensitivity analysis in the context of iris presentation attack detection (PAD). A presentation attack (PA) occurs when an adversary presents a fake or altered biometric sample such as printed eyes, plastic eyes, or cosmetic contact lenses to circumvent the iris recognition system \cite{ISO30107-1-2016}. Our application is to detect these PAs launched against an iris system. We formulate the detection problem as a two-class problem based on DNNs, where the input is a near-infrared iris image and the output is a PA score (range from 0-1) which is based on a specified threshold labeled as ``bonafide" or ``PA".

\section{Datasets and Experimental Setup}
\label{sec:DatasetSetup}

\begin{table*}[h]
\caption{Summary of training and test datasets along with the number of bonafide and PA iris images present in the datasets. The information about the sensors used to capture images is also provided. Here, “K. Test” means a known test set of the dataset, and “U. Test” means an unknown test set (see text for explanation).}
\label{table:Datasets}
\resizebox{\textwidth}{!}{%
\begin{tabular}{|l|clc|ccccccc|}
\hline
\textbf{Train/Test} & \multicolumn{3}{c|}{\textbf{Train}} & \multicolumn{7}{c|}{\textbf{Test}} \\ \hline
\textbf{Datasets} & \multicolumn{1}{c|}{\multirow{3}{*}{\textbf{IARPA}}} & \multicolumn{1}{l|}{\multirow{3}{*}{\textbf{\begin{tabular}[c]{@{}l@{}}NDCLD\\ -2015\end{tabular}}}} & \multicolumn{1}{l|}{\multirow{3}{*}{\textbf{\begin{tabular}[c]{@{}l@{}}Warsaw \\ PostMortem\\ v3\end{tabular}}}} & \multicolumn{6}{c|}{\textbf{LivDet-Iris-2017}} & \multicolumn{1}{l|}{\multirow{3}{*}{\textbf{LivDet-Iris-2020}}} \\ \cline{1-1} \cline{5-10}
\textbf{\begin{tabular}[c]{@{}l@{}}Dataset \\ Subsets\end{tabular}} & \multicolumn{1}{c|}{} & \multicolumn{1}{l|}{} & \multicolumn{1}{l|}{} & \multicolumn{1}{c|}{\textbf{\begin{tabular}[c]{@{}c@{}}Clarkson\\ (Cross-PA)\end{tabular}}} & \multicolumn{2}{c|}{\textbf{\begin{tabular}[c]{@{}c@{}}Warsaw\\ (Cross-sensor)\end{tabular}}} & \multicolumn{2}{c|}{\textbf{\begin{tabular}[c]{@{}c@{}}Notre Dame\\ (Cross-PA)\end{tabular}}} & \multicolumn{1}{c|}{\textbf{\begin{tabular}[c]{@{}c@{}}IIITD-WVU\\ (Cross-Dataset)\end{tabular}}} & \multicolumn{1}{l|}{} \\ \cline{1-1} \cline{5-10}
\textbf{Splits} & \multicolumn{1}{c|}{} & \multicolumn{1}{l|}{} & \multicolumn{1}{l|}{} & \multicolumn{1}{c|}{\textbf{Test}} & \multicolumn{1}{c|}{\textbf{K. Test}} & \multicolumn{1}{c|}{\textbf{U. Test}} & \multicolumn{1}{c|}{\textbf{K. Test}} & \multicolumn{1}{c|}{\textbf{U. Test}} & \multicolumn{1}{c|}{\textbf{Test}} & \multicolumn{1}{l|}{} \\ \hline
Bonafide & \multicolumn{1}{c|}{19,453} & \multicolumn{1}{c|}{-} & - & \multicolumn{1}{c|}{1,485} & \multicolumn{1}{c|}{974} & \multicolumn{1}{c|}{2,350} & \multicolumn{1}{c|}{900} & \multicolumn{1}{c|}{900} & \multicolumn{1}{c|}{702} & 5,331 \\ \hline
Print & \multicolumn{1}{c|}{1,005} & \multicolumn{1}{c|}{-} & - & \multicolumn{1}{c|}{908} & \multicolumn{1}{c|}{2,016} & \multicolumn{1}{c|}{2,160} & \multicolumn{1}{c|}{-} & \multicolumn{1}{c|}{-} & \multicolumn{1}{c|}{2,806} & 1,049 \\ \hline
\begin{tabular}[c]{@{}l@{}}Cosmetic \\ Contacts\end{tabular} & \multicolumn{1}{c|}{1,187} & \multicolumn{1}{l|}{2,236} & - & \multicolumn{1}{c|}{765} & \multicolumn{1}{c|}{-} & \multicolumn{1}{c|}{-} & \multicolumn{1}{c|}{900} & \multicolumn{1}{c|}{900} & \multicolumn{1}{c|}{701} & 4,336 \\ \hline
\begin{tabular}[c]{@{}l@{}}Artificial\\ Eyes\end{tabular} & \multicolumn{1}{c|}{1,804} & \multicolumn{1}{c|}{-} & - & \multicolumn{1}{c|}{-} & \multicolumn{1}{c|}{-} & \multicolumn{1}{c|}{-} & \multicolumn{1}{c|}{-} & \multicolumn{1}{c|}{-} & \multicolumn{1}{c|}{-} & 541 \\ \hline
\begin{tabular}[c]{@{}l@{}}Electronic \\ Display\end{tabular} & \multicolumn{1}{c|}{51} & \multicolumn{1}{c|}{-} & - & \multicolumn{1}{c|}{-} & \multicolumn{1}{c|}{-} & \multicolumn{1}{c|}{-} & \multicolumn{1}{c|}{-} & \multicolumn{1}{c|}{-} & \multicolumn{1}{c|}{-} & 81 \\ \hline
Cadaver Eyes & \multicolumn{1}{c|}{-} & \multicolumn{1}{c|}{-} & \multicolumn{1}{l|}{1,200} & \multicolumn{1}{c|}{-} & \multicolumn{1}{c|}{-} & \multicolumn{1}{c|}{-} & \multicolumn{1}{c|}{-} & \multicolumn{1}{c|}{-} & \multicolumn{1}{c|}{-} & 1,094 \\ \hline
Sensor & \multicolumn{1}{c|}{\begin{tabular}[c]{@{}c@{}}COTS Iris\\ Sensors x3$^1$\end{tabular}} & \multicolumn{1}{l|}{\begin{tabular}[c]{@{}l@{}}IrisGuard\\ AD100,\\ IrisAccess\\ LG4000\end{tabular}} & \begin{tabular}[c]{@{}c@{}}IriShield \\ MK2120U\end{tabular} & \multicolumn{1}{c|}{\begin{tabular}[c]{@{}c@{}}IrisAccess\\ EOU2200\end{tabular}} & \multicolumn{1}{c|}{\begin{tabular}[c]{@{}c@{}}IrisGuard\\ AD100\end{tabular}} & \multicolumn{1}{c|}{\begin{tabular}[c]{@{}c@{}}Aritech ARX-3M3C,\\ Fujinon DV10X7.5A,\\ DV10X7.5A-SA2 lens\\ B+W 092 NIR filter\end{tabular}} & \multicolumn{2}{c|}{\begin{tabular}[c]{@{}c@{}}IrisGuard AD100,\\ IrisAccess LG4000\end{tabular}} & \multicolumn{1}{c|}{\begin{tabular}[c]{@{}c@{}}IriShield \\ MK2120U\end{tabular}} & \multicolumn{1}{l|}{\begin{tabular}[c]{@{}l@{}}Iris ID iCAM7000,\\ IrisGuardAD100,\\ IrisAccess LG4000,\\ IriTech IriShield\end{tabular}} \\ \hline
\multicolumn{8}{l}{$^1$Specific sensor names withheld at sponsor’s request}
\end{tabular}
 }
\end{table*}


The training data we use to build our iris PAD models are IARPA, NDCLD-2015 \cite{NDCLD2015} and Warsaw PostMortem v3 \cite{Trokielewicz2020} datasets. The IARPA dataset is a proprietary dataset consisting of 19,453 bonafide irides and 4,047 presentation attack (PA) samples. From the NDCLD-2015 dataset, we use 2,236 cosmetic contact lens images for training. From the Warsaw PostMortem v3 dataset, 1,200 cadaver iris images from the first 37 cadavers are used for training. Testing is performed on the LivDet-Iris-2017 \cite{Yambay2017} and LivDet-Iris-2020 \cite{Das2020} datasets. Both of these are publicly available competition datasets for evaluating iris presentation attack detection performance. The LivDet-Iris-2017 dataset \cite{Yambay2017} consists of four subsets: Clarkson, Warsaw, Notre Dame, and IIITD-WVU. All subsets contain train and test partitions, and we use only the test partition. Warsaw and Notre Dame subsets further contains two splits in the test partition: `Known' and `Unknown'. The `Known' split corresponds to the scenario where PAs of the same type or images from similar sensors are present in both train and test partitions, while the `Unknown' split contains different types of PAs or images from different types of sensors in the train and test partitions. In our case, both test splits are considered as `Unknown' type as we use different datasets for training. Such a testing scenario is referred to cross-dataset. However, we keep the original terminologies (`Known' and `Unknown') of test splits in the work. 
The LivDet-Iris-2020 \cite{Das2020} consists of a single test split, and this scenario also corresponds to cross-dataset. Table \ref{table:Datasets} describes all training and test sets, along with the types of PAs and images present in them. In aggregate, both datasets provide a diverse set of PAs. 
 
We use three iris PA detectors for sensitivity analysis. Two of the detectors utilize VGG19 \cite{Simonyan2015} and ResNet101 \cite{He2016} networks as their backbone architecture. The third detector is D-NetPAD \cite{Sharma2020}, where the backbone architecture is DenseNet161 \cite{Huang2017}. The D-NetPAD shows state-of-the-art performance on both LivDet-Iris-2017 and LivDet-Iris-2020 iris PAD competitions \cite{Sharma2020,Das2020}. Since D-NetPAD already had the state-of-the-art performance on the evaluation datasets and Smith \emph{et. al.} \cite{smith2023convnets} found that convolution-based networks can perform same as vision transformer at scale, we did not perform similar analysis or experiments on transformer-based models like ViT \cite{dosovitskiy2020image}. The convolutional networks we use require a cropped iris region resized to 224 $\times$ 224 as input. For training, we initialize the model with the weights from the ImageNet dataset \cite{Deng2009} and then fine-tune the models using the training datasets described above. The learning rate was set to 0.005, the batch size was 20, the number of epochs was 50, the optimization algorithm was stochastic gradient descent with a momentum of 0.9, and the loss function used was cross-entropy. 

We measure the sensitivity of these DNNs by evaluating their performance as a function of the weight perturbations. The performance is estimated in terms of TDR (\%) at 0.2\% False Detection Rate (FDR).\footnote{The threshold at this specific FDR was selected by the sponsor.}
FDR is the percentage of bonafide samples incorrectly classified as PAs.\footnote{ISO/IEC 30107-3:2023 specifies Attack Presentation Classification Error Rate (APCER) and Bonafide Presentation Classification Error Rate (BPCER) as evaluation metrics for PAD. TDR is 1$-$APCER, and FDR is the same as BPCER.} In Table \ref{table:Fusion-Results}, the row corresponding to the `Original' method reports the performance of these models on the LivDet-Iris-2017 and LivDet-Iris-2020 datasets {\em before} weights were perturbed. On the LivDet-Iris-2017 dataset, ResNet101 performs the best (average 74.55\% TDR), whereas on the LivDet-Iris-2020 dataset, D-NetPAD performs the best (90.22\% TDR). We also provide information about the number of weights and bias parameters present in all three models (Table \ref{table:Parameters}). The VGG19 architecture has the highest number of parameters, followed by the ResNet101 architecture.


\begin{table}[]
\caption{The number of parameters (weights and bias) present in all convolutional layers and the entire network of the VGG19, ResNet101, and D-NetPAD architectures.}
\label{table:Parameters}
\resizebox{\columnwidth}{!}{%
\begin{tabular}{|l|l|l|l|}
\hline
\textbf{Architecture} & \textbf{VGG19} & \textbf{ResNet101} & \textbf{D-NetPAD} \\ \hline
Weights & 139,570,240 & 42,451,584 & 26,366,448 \\ \hline
Bias & 19,202 & 52,674 & 109,970 \\ \hline
Total & 139,589,442 & 42,504,258 & 26,476,418 \\ \hline
\end{tabular}
}
\end{table}

\section{Sensitivity Analysis}
\label{sec:SensitivityAnalysis}

\subsection{Gaussian Noise Addition}

\begin{figure*}[t]
     \centering
     \begin{subfigure}[b]{0.49\textwidth}
         \centering
         \includegraphics[width=0.99\textwidth]{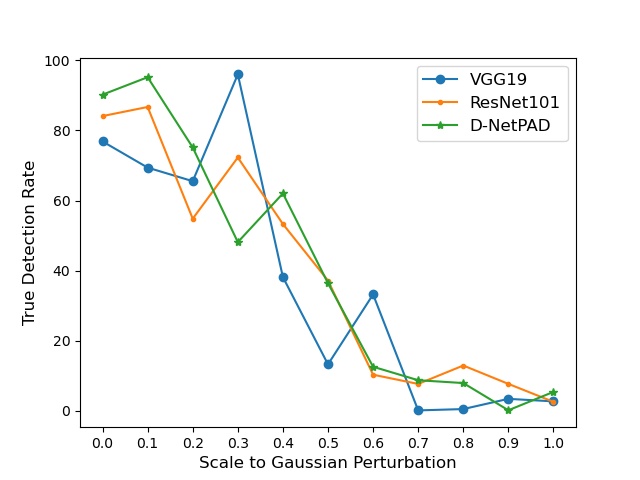}
         \caption{}
         \label{fig:Gaussian-Manipulations-Entire}
     \end{subfigure}
     \hfill
     \begin{subfigure}[b]{0.49\textwidth}
         \centering
         \includegraphics[width=0.99\textwidth]{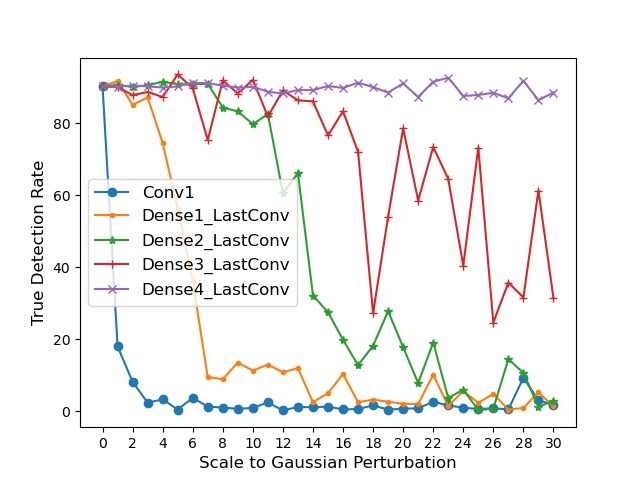}
         \caption{}
         \label{fig:Gaussian-Manipulations-Layers}
     \end{subfigure}
        \caption{Gaussian noise manipulation: (a) Performance (TDR at 0.2\% FDR) of VGG19, ResNet101, and D-NetPAD when weights and bias parameters of the entire network are perturbed. (b) Performance of D-NetPAD when the individual layer’s parameters (weights and bias) are perturbed. Here, Conv1 means the first convolution layer of the D-NetPAD, Dense1\textunderscore LastConv means the last convolution layer of the first dense block, and so on.} 
\end{figure*}

The Gaussian noise manipulation involves the addition of Gaussian noise to the original parameters. Figure \ref{fig:Gaussian-Manipulations-Entire} shows the performance of all the networks when we perturb parameters of all layers with the Gaussian noise. The scale factor ($\boldsymbol{\alpha}$) used to modify the standard deviation is shown on the x-axis. Every data point in the figure represents a single performance of the model. From a trend standpoint, the performance of all networks decreases with an increase in the standard deviation. However, this decrease is not linear. In fact, there are some performance gains at certain scales. These scales are different for different networks. For instance, the VGG19 network shows improvement for $\boldsymbol{\alpha}=$ 0.3, 0.6, and 0.9, ResNet101 for $\boldsymbol{\alpha}=$ 0.1, 0.3, and 0.9, and D-NetPAD for $\boldsymbol{\alpha}=$ 0.1, 0.4 and 1.0. Surprisingly, certain scales give higher performance than the original model, such as 0.1 scale for the ResNet101 and D-NetPAD models, and 0.3 scale for the VGG19 model. The results indicate that all three networks are sensitive to Gaussian noise perturbations when perturbations are applied over all layers of the network, and we cannot conclude which network is comparatively stable under these weight perturbations. 


We further analyze the impact of perturbation at different layers on the performance of the models. We manipulate the parameters one layer at a time and observe the performance change. For the layer-wise analysis, we show the results for only the D-NetPAD model since the other two models also show similar performance trends. In the case of D-NetPAD, we select the first convolution layer and the last convolution layers of four dense blocks for perturbation. Figure \ref{fig:Gaussian-Manipulations-Layers} shows the performance of D-NetPAD when the individual layer’s parameters are perturbed. We observe that the initial layers have more influence on the performance of the D-NetPAD compared to the later layers. The model is highly robust to the perturbations in the last convolution layer of the fourth dense block, even at a scale factor of 30. Cheney \textit{et al.} \cite{Cheney2017} also observe the higher impact of perturbations in the initial layers on the performance. Generally, initial layers focus more on capturing discriminative or representative features, whereas later layers are more responsible for generating decision boundaries. Manipulations to extracted features have more impact on the performance compared to a slight change in decision boundaries. Moreover, manipulation in initial layers changes feature maps of all subsequent layers and, hence, causes propagation of error. Change in middle layers exhibit large fluctuations in performance compared to the initial and later layers. 

\subsection{Weight Zeroing}
\begin{figure*}[h]
     \centering
     \begin{subfigure}[b]{0.49\textwidth}
         \centering
         \includegraphics[width=0.99\textwidth]{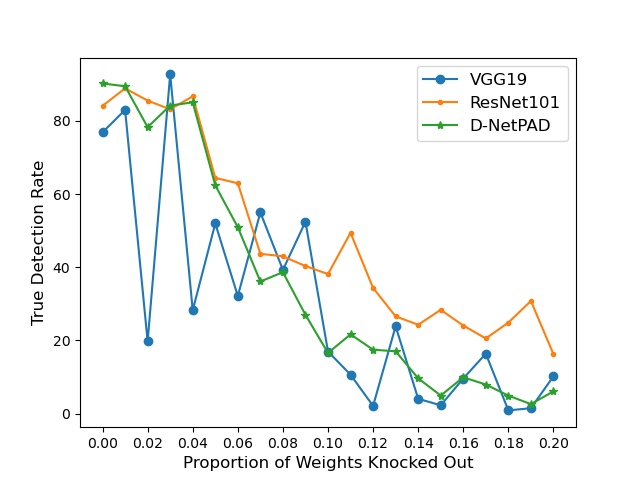}
         \caption{}
         \label{fig:WeightZeroPerturbation-Entire}
     \end{subfigure}
     \hfill
     \begin{subfigure}[b]{0.49\textwidth}
         \centering
         \includegraphics[width=0.99\textwidth]{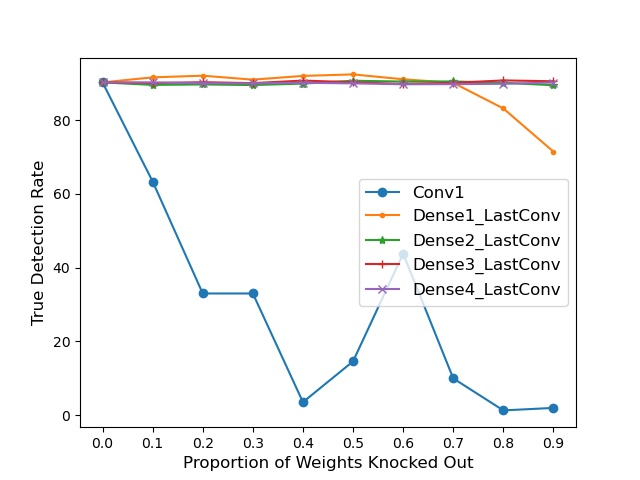}
         \caption{}
         \label{fig:WeightZeroPerturbation-Layers}
     \end{subfigure}
        \caption{Weight zeroing manipulation: (a) Performance (TDR at 0.2\% FDR) of VGG19, ResNet101, and D-NetPAD when parameters of the entire network are perturbed. (b) Performance of D-NetPAD when the individual layer’s parameters are perturbed.} 
\end{figure*}

The weight zeroing manipulation involves random selection of a particular fraction of weight parameters and setting them to zero. Figure \ref{fig:WeightZeroPerturbation-Entire} shows the performance of all three architectures when we manipulate the entire set of network parameters, while Figure \ref{fig:WeightZeroPerturbation-Layers} shows the performance of D-NetPAD when we perturb individual layers. Similar conclusions can be drawn from Figure \ref{fig:WeightZeroPerturbation-Entire} as drawn from Figure \ref{fig:Gaussian-Manipulations-Entire} that the overall performance of all three architectures decreases with an increase in the proportion of weights set to zero. However, certain perturbations give improved performance. For example zeroing 3\% of weights improves the VGG19 network performance from 76.87\% TDR (original) to 92.70\% TDR. In the case of ResNet101, zeroing 3\% of weights improves performance from 84.11\% TDR (original) to 88.88\% TDR. Again, all three networks are sensitive to the zeroing out of randomly selected weights.

\begin{figure*}
    \centering
    \begin{subfigure}[b]{0.18\textwidth}
         \centering
         \includegraphics[width=0.99\textwidth]{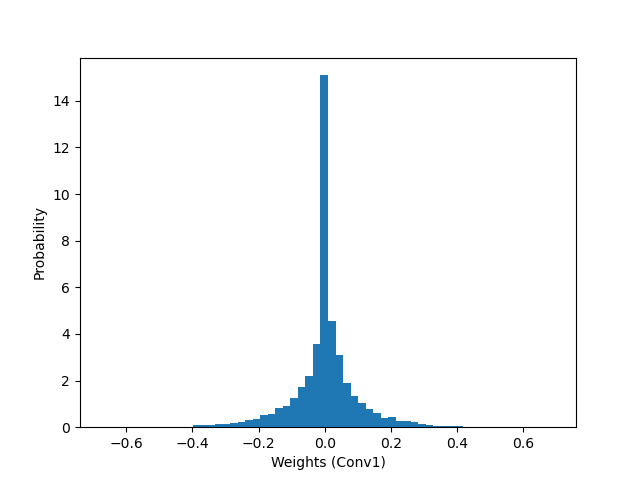}
         \caption{Conv1 \\$\mu$: 0.0, $\sigma$: 0.103}
     \end{subfigure}
     \begin{subfigure}[b]{0.18\textwidth}
         \centering
         \includegraphics[width=0.99\textwidth]{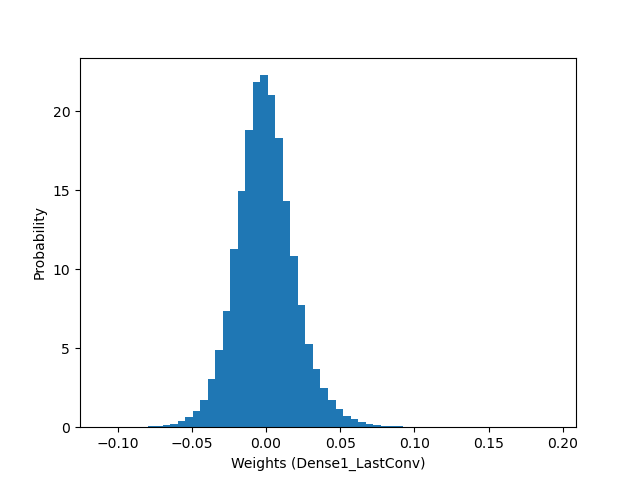}
         \caption{Dense1\textunderscore LastConv \\ $\mu$: 0.0, $\sigma$: 0.019}
     \end{subfigure}
     \begin{subfigure}[b]{0.18\textwidth}
         \centering
         \includegraphics[width=0.99\textwidth]{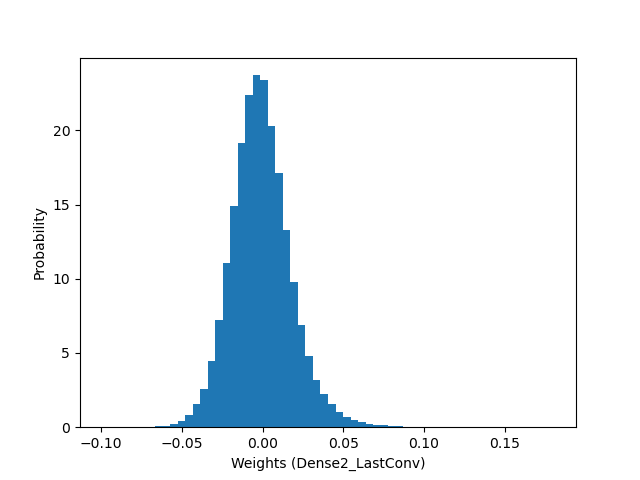}
         \caption{Dense2\textunderscore LastConv \\ $\mu$: 0.0, $\sigma$: 0.018}
     \end{subfigure}
     \begin{subfigure}[b]{0.18\textwidth}
         \centering
         \includegraphics[width=0.99\textwidth]{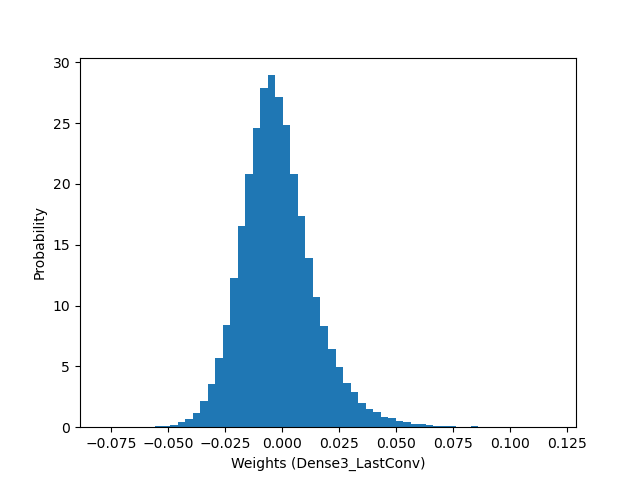}
         \caption{Dense3\textunderscore LastConv \\ $\mu$: 0.0, $\sigma$: 0.016}
     \end{subfigure}
     \begin{subfigure}[b]{0.18\textwidth}
         \centering
         \includegraphics[width=0.99\textwidth]{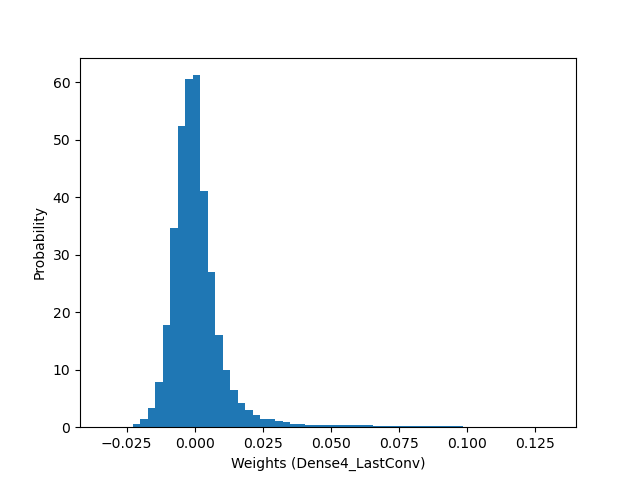}
         \caption{Dense4\textunderscore LastConv \\$\mu$: -0.0, $\sigma$: 0.012}
     \end{subfigure}
    \caption{Weight distribution of different layers of the trained D-NetPAD architecture. Mean ($\mu$) and standard deviation ($\sigma$) are provided below each distribution.}
    \label{fig:Weight_Distributions}
\end{figure*}

In the layer-wise setup (Figure \ref{fig:WeightZeroPerturbation-Layers}), the performance of D-NetPAD is stable except for the first convolution layer. This is due to the fact that the original weights of the convolution layers have a zero mean and a small standard deviation ranging from 0.10 (first convolution layer) to 0.01 (last convolution layer) as shown in Figure \ref{fig:Weight_Distributions}. Initial layers have a higher standard deviation compared to later layers, which makes the network more sensitive to the manipulations in the initial layers. A similar performance trend is observed in the VGG19 and ResNet101 networks as well. 

\begin{figure*}[h]
     \centering
     \begin{subfigure}[b]{0.49\textwidth}
         \centering
         \includegraphics[width=0.99\textwidth]{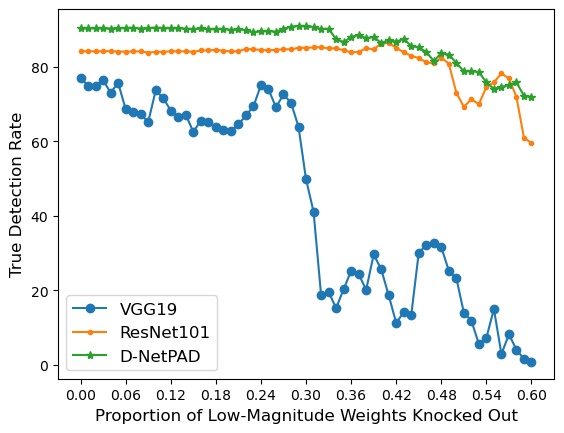}
         \caption{}
         \label{fig:BottomWeightZeroPerturbation-Entire}
     \end{subfigure}
     \hfill
     \begin{subfigure}[b]{0.49\textwidth}
         \centering
         \includegraphics[width=0.99\textwidth]{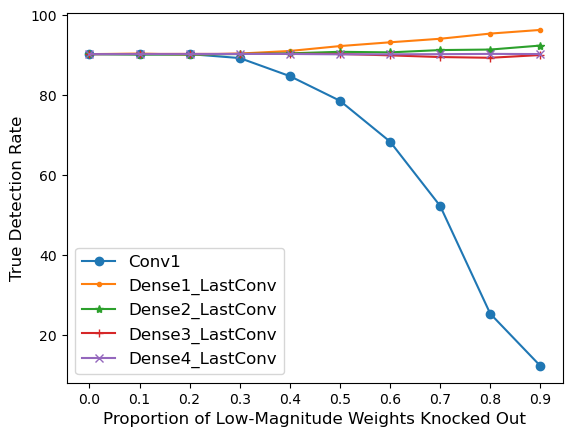}
         \caption{}
         \label{fig:BottomWeightZeroPerturbation-Layers}
     \end{subfigure}
        \caption{Variant of the weight zeroing manipulation (low-magnitude weights are set to zero): (a) Performance (TDR at 0.2\% FDR) of VGG19, ResNet101, and D-NetPAD when parameters of the entire network are perturbed. (b) Performance of D-NetPAD when individual layer’s parameters are perturbed.} 
\end{figure*}

Since most of the original weights are already close to 0, we apply a variant of weight zeroing where only low-magnitude weights are set to zero. Figure \ref{fig:BottomWeightZeroPerturbation-Entire} shows the performance of all architectures when we manipulate the entire network in this fashion, while Figure \ref{fig:BottomWeightZeroPerturbation-Layers} shows the performance of D-NetPAD on layer-wise manipulation. ResNet101 and D-NetPAD networks are observed to be robust to this manipulation as zeroing out even 33\% of all weights does not affect their performance. VGG19 also shows robustness with only a 6\% drop in performance, though its performance is not as stable as the ResNet101 and D-NetPAD networks. Figure \ref{fig:BottomWeightZeroPerturbation-Layers} shows the sensitivity of the D-NetPAD on layer-wise perturbations. Zeroing out even 30\% of the first convolution layer weights does not impact its performance. Remarkably, the manipulation in the last convolution layer of the first and second dense blocks shows a linear increase in performance. The performance of D-NetPAD increases from 90.22\% TDR to 96.28\% TDR upon manipulating the last convolution layer of the first dense block. This implies that we could zero out low-magnitude weights and reduce the size of the model without affecting its performance. This finding is useful in building a compressed DNN model with better time and memory efficiency to deploy on mobile or embedded devices.


\subsection{Weight Scaling}
\begin{figure*}
     \centering
     \begin{subfigure}[b]{0.49\textwidth}
         \centering
         \includegraphics[width=0.99\textwidth]{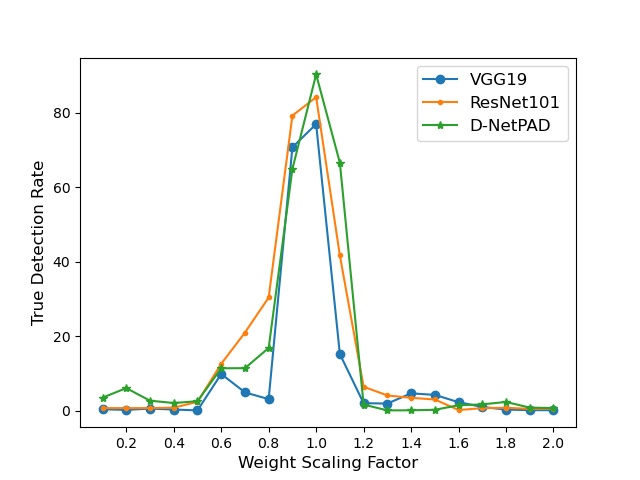}
         \caption{}
         \label{fig:WeightScalingPerturbation-Entire}
     \end{subfigure}
     \hfill
     \begin{subfigure}[b]{0.49\textwidth}
         \centering
         \includegraphics[width=0.99\textwidth]{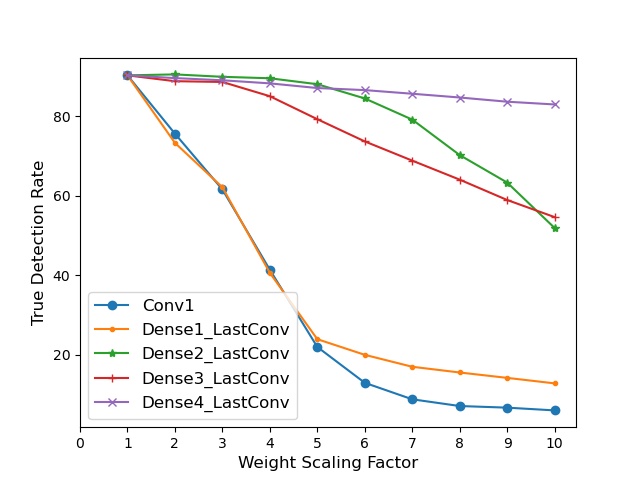}
         \caption{}
         \label{fig:WeightScalingPerturbation-Layers}
     \end{subfigure}
        \caption{Weight scaling manipulation: (a) Performance (TDR at 0.2\% FDR) of VGG19, ResNet101, and D-NetPAD when parameters of the entire network are perturbed  simultaneously. (b) Performance of D-NetPAD when the individual layer’s parameters are perturbed.} 
\end{figure*}

This manipulation scales the original parameters with a scalar value. Figure \ref{fig:WeightScalingPerturbation-Entire} shows the performance of all three architectures when we manipulate the entire set of network parameters, while Figure \ref{fig:WeightScalingPerturbation-Layers} presents the performance of D-NetPAD when we perturb specific layers. The performance at scale 1 indicates the original performance without weight perturbations. Weight perturbations across the entire network resulted in a radical drop in performance even with a small scalar factor (0.8 or 1.1). In the layer-wise manipulation, the initial layers show a higher impact on the performance of D-NetPAD compared to the later layers. The manipulation in the last convolution layer does not impact the performance even at a scaling factor of 10. A similar performance trend is observed on the VGG19 and ResNet101 networks as well.

\subsection{Findings}
Here are the main findings from the aforementioned analysis:

\noindent 1. All three networks decrease in performance when perturbations are applied over the entire network. \footnote{The notion of `significant' change in performance is a subject of future work.} However, the networks show robustness when low-magnitude weights are set to zero. The scaling of weights has a major negative impact on the performance of networks. 

\noindent 2. Layer-wise sensitivity analysis shows that perturbations in initial layers impacted the performance to a greater extent compared to the later layers.
The weight distribution of all layers are zero-centered and later layers have a lower standard deviation compared to initial layers (Figure \ref{fig:Weight_Distributions}), making later layers less sensitive to weight zeroing and scaling perturbations as majority of their weights are already close to the zero mean. The zero-centered nature of weight distributions is also a reason why Gaussian noise perturbations have the most negative impact on the performance compared to the other perturbations.

\noindent 3. Certain perturbations improve the performance of network models over the original one in both settings (entire network and layer-wise). This observation indicates that the parameters learned by the models during training are not optimum. Random change in the weights in their close vicinity shows improvement in the performance. Hence, there is further scope for optimizing weights.

\noindent 4. Zeroing out low-magnitude weights results in better performance as well as reduces the size of the model.    

\begin{table*}[]
\caption{The performance of VGG19, ResNet101, and D-NetPAD models in terms of True Detection Rate (\%, higher the better) at 0.2\% False Detection Rate on the LivDet-Iris-2017 and LivDet-Iris-2020 datasets. The performance is shown on original model (no parameter perturbations), perturbed model and an ensemble of model.}
\label{table:Fusion-Results}
\centering
\begin{tabular}{|lccccccc|}
\hline
\multicolumn{1}{|l|}{\textbf{Datasets}} & \multicolumn{6}{c|}{\textbf{LivDet-Iris-2017}} & \multicolumn{1}{l|}{\multirow{3}{*}{\textbf{LivDet-Iris-2020}}} \\ \cline{1-7}
\multicolumn{1}{|l|}{\textbf{Subsets}} & \multicolumn{1}{c|}{\textbf{Clarkson}} & \multicolumn{2}{c|}{\textbf{Warsaw}} & \multicolumn{2}{c|}{\textbf{Notre Dame}} & \multicolumn{1}{c|}{\textbf{IIITD-WVU}} & \multicolumn{1}{l|}{} \\ \cline{1-7}
\multicolumn{1}{|l|}{\textbf{Splits}} & \multicolumn{1}{c|}{\textbf{Test}} & \multicolumn{1}{c|}{\textbf{K. Test}} & \multicolumn{1}{c|}{\textbf{U. Test}} & \multicolumn{1}{c|}{\textbf{K. Test}} & \multicolumn{1}{c|}{\textbf{U. Test}} & \multicolumn{1}{c|}{\textbf{Test}} & \multicolumn{1}{l|}{} \\ \hline
\multicolumn{8}{|c|}{\textbf{VGG19 Model}} \\ \hline
\multicolumn{1}{|l|}{Original} & \multicolumn{1}{c|}{51.32} & \multicolumn{1}{c|}{86.25} & \multicolumn{1}{c|}{10.12} & \multicolumn{1}{c|}{\textbf{100}} & \multicolumn{1}{c|}{\textbf{99.00}} & \multicolumn{1}{c|}{1.44} & 76.87 \\ \hline
\multicolumn{1}{|l|}{Perturbed} & \multicolumn{1}{c|}{54.88} & \multicolumn{1}{c|}{91.12} & \multicolumn{1}{c|}{9.08} & \multicolumn{1}{c|}{\textbf{100}} & \multicolumn{1}{c|}{97.78} & \multicolumn{1}{c|}{1.58} & \textbf{90.31} \\ \hline
\multicolumn{1}{|l|}{Ensemble (Score-level)} & \multicolumn{1}{c|}{66.17} & \multicolumn{1}{c|}{\textbf{92.95}} & \multicolumn{1}{c|}{7.23} & \multicolumn{1}{c|}{\textbf{100}} & \multicolumn{1}{c|}{98.00} & \multicolumn{1}{c|}{3.14} & 89.53 \\ \hline
\multicolumn{1}{|l|}{Ensemble (Parameter-level)} & \multicolumn{1}{c|}{\textbf{73.01}} & \multicolumn{1}{c|}{84.92} & \multicolumn{1}{c|}{\textbf{13.90}} & \multicolumn{1}{c|}{99.78} & \multicolumn{1}{c|}{97.78} & \multicolumn{1}{c|}{\textbf{9.43}} & 88.26 \\ \hline
\multicolumn{8}{|c|}{\textbf{ResNet101 Model}} \\ \hline
\multicolumn{1}{|l|}{Original} & \multicolumn{1}{c|}{15.82} & \multicolumn{1}{c|}{89.93} & \multicolumn{1}{c|}{91.67} & \multicolumn{1}{c|}{\textbf{100}} & \multicolumn{1}{c|}{\textbf{99.44}} & \multicolumn{1}{c|}{50.47} & 84.11 \\ \hline
\multicolumn{1}{|l|}{Perturbed} & \multicolumn{1}{c|}{\textbf{23.01}} & \multicolumn{1}{c|}{\textbf{95.33}} & \multicolumn{1}{c|}{\textbf{94.65}} & \multicolumn{1}{c|}{\textbf{100}} & \multicolumn{1}{c|}{95.67} & \multicolumn{1}{c|}{\textbf{58.14}} & 86.40 \\ \hline
\multicolumn{1}{|l|}{Ensemble (Score-level)} & \multicolumn{1}{c|}{21.61} & \multicolumn{1}{c|}{92.95} & \multicolumn{1}{c|}{94.60} & \multicolumn{1}{c|}{\textbf{100}} & \multicolumn{1}{c|}{89.88} & \multicolumn{1}{c|}{58.02} & \textbf{91.07} \\ \hline
\multicolumn{1}{|l|}{Ensemble (Parameter-level)} & \multicolumn{1}{c|}{19.10} & \multicolumn{1}{c|}{\textbf{95.33}} & \multicolumn{1}{c|}{94.37} & \multicolumn{1}{c|}{\textbf{100}} & \multicolumn{1}{c|}{95.67} & \multicolumn{1}{c|}{\textbf{58.14}} & 89.92 \\ \hline
\multicolumn{8}{|c|}{\textbf{D-NetPAD Model}} \\ \hline
\multicolumn{1}{|l|}{Original} & \multicolumn{1}{c|}{60.04} & \multicolumn{1}{c|}{76.68} & \multicolumn{1}{c|}{35.76} & \multicolumn{1}{c|}{\textbf{100}} & \multicolumn{1}{c|}{\textbf{99.33}} & \multicolumn{1}{c|}{32.01} & 90.22 \\ \hline
\multicolumn{1}{|l|}{Perturbed} & \multicolumn{1}{c|}{\textbf{68.54}} & \multicolumn{1}{c|}{\textbf{94.94}} & \multicolumn{1}{c|}{\textbf{53.02}} & \multicolumn{1}{c|}{\textbf{100}} & \multicolumn{1}{c|}{99.00} & \multicolumn{1}{c|}{\textbf{50.35}} & \textbf{96.86} \\ \hline
\multicolumn{1}{|l|}{Ensemble (Score-level)} & \multicolumn{1}{c|}{68.34} & \multicolumn{1}{c|}{93.84} & \multicolumn{1}{c|}{46.40} & \multicolumn{1}{c|}{\textbf{100}} & \multicolumn{1}{c|}{97.66} & \multicolumn{1}{c|}{48.08} & 96.71 \\ \hline
\multicolumn{1}{|l|}{Ensemble (Parameter-level)} & \multicolumn{1}{c|}{64.29} & \multicolumn{1}{c|}{\textbf{94.94}} & \multicolumn{1}{c|}{\textbf{53.02}} & \multicolumn{1}{c|}{\textbf{100}} & \multicolumn{1}{c|}{99.00} & \multicolumn{1}{c|}{42.59} & 95.66 \\ \hline
\end{tabular}
\end{table*}

\section{Performance Improvement}
\label{sec:ProposedImprovement}
We observe that certain perturbations result in better performance, even higher than that of the original model. We leverage this observation and obtain better performing models using these perturbations without any additional training. In this regard, we explore two directions: the first is to find a single perturbed model which achieves good performance consistently, and the second is to create an ensemble of high-performing perturbed models. In the earlier part of the work, we analyzed the sensitivity of different architectures based on their performance on the LivDet-Iris-2020 dataset. Here, we select a high-performing perturbed model and validate its performance on the LivDet-Iris-2017 dataset. For the ensemble of models, we further explore two sub-directions based on the level of fusion. In the first, we simply fuse their decision scores using the sum rule. This level of fusion better spans the decision space and generalizes well to the test data \cite{Polikar2006}. However, it increases the inference time as decision scores are required from all the component models. The second level of fusion is at the model parameter-level, where we fuse the parameters by averaging and merge the component models into one model. This level of fusion better spans the parameter space and reduces the inference time as a decision is required from only one model.
Details of these models are given below:

    \noindent \textbf{1. Original Model:} The model utilizes originally trained parameters without any perturbation of the parameters.
    
    \noindent \textbf{2. Perturbed Model:} 
     In the case of VGG19, we create a perturbed model by setting 95\% of the low-magnitude weights of the seventh convolution layer to zero. For the ResNet101 model, a perturbed model is formed by setting 40\% of low-magnitude weights of the first convolution layer to zero, while for the D-NetPAD, 92\% of the low-magnitude weights of the last convolution layer of the first dense block are set to zero. The selection of these perturbed models are based on their consistent high performance on the LivDet-Iris-2020 dataset (\ref{fig:BottomWeightZeroPerturbation-Layers}). We repeat the experiment 100 times for each of the high-performing models and select the one with the most consistent performance.
    
    
    \noindent \textbf{3. Ensemble Models at the Score-Level:} We combine two consistent high-performing perturbed models by fusing their PA scores using the sum rule. For all three architectures, we fuse the above specified perturbed models with the model formed by adding Gaussian noise with $\boldsymbol{\alpha} = 0.1$ ($N(0,0.3*\sigma(W_{org})$) to the entire network.

    \noindent \textbf{4. Ensemble Models at the Parameter-Level:} We create a single ensemble model by averaging the parameters of two consistent high-performing perturbed models. The PA score is generated from a single merged model. The models selected for fusion are the same ones used for ensembling at the score-level.
    

Table \ref{table:Fusion-Results} provides the performance of these models (based on VGG19, ResNet101, and D-NetPAD architectures). The performance of perturbed and ensemble models is better than the original model on both datasets. The observation holds true for all three architectures. The perturbed models show an average improvement of 47.12\% and 8.97\%, the ensemble model at the score-level shows an improvement of 16.01\% and 10.65\%, and the ensemble model at the parameter-level shows an improvement of 43.58\% and 9.25\% on the LivDet-Iris-2017 and LivDet-Iris-2020 datasets, respectively. One major advantage of these perturbed models is that these models are created without any further training. Another advantage is that these high-performing perturbed models have reduced model size.

\section{Summary and Future Work}
\label{sec:Summary}
We analyze the sensitivity of three DNN architectures (VGG19, ResNet101, and D-NetPAD) under three types of parameter perturbations (Gaussian noise manipulation, weight zeroing, and weight scaling). We apply the perturbations in two settings: modifying the weights across all layers and modifying weights layer-by-layer. We found that CNNs are generally less sensitive to a variant of weight zeroing, where low-magnitude weights are set to zero. From the layer-wise analysis, we observe that the CNNs are more robust to perturbations in later layers compared to the initial layers and Gaussian noise addition most negatively impacts the performance due to the zero-centered nature of weight distributions. Certain manipulations improve the performance over the original one. Based on these observations, we propose the use of an ensemble of models that consistently perform well on both LivDet-Iris-2017 and LivDet-Iris-2020 datasets. As future work, we will focus on finding the analytical optimum direction for weight perturbations. Additionally, the approach can be applied to other domains and tasks.

\section*{Acknowledgments}
This research is based upon work supported in part by the Office of the Director of National Intelligence (ODNI), Intelligence Advanced Research Projects Activity (IARPA), via IARPA R\&D Contract No. 2017 - 17020200004. The views and conclusions contained herein are those of the authors and should not be interpreted as necessarily representing the official policies, either expressed or implied, of ODNI, IARPA, or the U.S. Government. The U.S. Government is authorized to reproduce and distribute reprints for governmental purposes notwithstanding any copyright annotation therein.

\balance
{\small
\bibliographystyle{ieee_fullname}
\bibliography{main}
}

\end{document}